\documentclass[conference]{IEEEtran}
\IEEEoverridecommandlockouts
\usepackage{cite}
\usepackage{amsmath,amssymb,amsfonts}
\usepackage{algorithmic}
\usepackage{algorithm}
\usepackage{graphicx}
\usepackage{textcomp}
\usepackage{xcolor}
\usepackage{booktabs}
\usepackage{multirow}
\usepackage{hyperref}
\usepackage{microtype}
\usepackage{url}
\usepackage{tikz}
\usepackage{pgfplots}
\pgfplotsset{compat=1.18}
\usetikzlibrary{arrows.meta,positioning,shapes.geometric}
\def\BibTeX{{\rm B\kern-.05em{\sc i\kern-.025em b}\kern-.08em
    T\kern-.1667em\lower.7ex\hbox{E}\kern-.125emX}}

\newcommand{\sysname}{CLARITY}

\begin{document}

\title{RAG-Based Auto-Configuration for Industrial Fieldbus Devices}

\author{
    \IEEEauthorblockN{%
        Aadil Gani Ganie\IEEEauthorrefmark{1},
        Saad Ezzini\IEEEauthorrefmark{1}\textsuperscript{\S}, and
        Naveed Farooz Marazi\IEEEauthorrefmark{2}%
    }
    \IEEEauthorblockA{%
        \IEEEauthorrefmark{1}Interdisciplinary Research Center for Intelligent Manufacturing and Robotics (IRC-IMR), KFUPM, Saudi Arabia\\
        \textsuperscript{\S}Department of Information and Computer Science, KFUPM, Saudi Arabia\\
        \IEEEauthorrefmark{2}Interdisciplinary Research Center for Smart Mobility and Logistics, KFUPM, Saudi Arabia\\
        Email: \{aadil.ganie, saad.ezzini, naveed.marazi\}@kfupm.edu.sa
    }
}

\maketitle

% ============================================================
\begin{abstract}
Industrial device commissioning demands that automation engineers
manually extract hundreds of protocol-specific parameters from
heterogeneous PDF manuals and transcribe them into supervisory
control systems, a workflow that is time-intensive and prone to
transcription errors. This paper presents \sysname{}
, a production-oriented pipeline that automates
device configuration end-to-end for Modbus RTU, OPC-UA, Profibus
DP, and CANopen. \sysname{} builds a hybrid dense-sparse
retrieval index augmented by an ontology graph derived from
ECLASS, Asset Administration Shell (AAS), and SOSA/SSN, and uses
a 1024-dimensional BGE-M3 encoder with a cross-encoder reranker
to surface relevant manual passages. A local LLM ($T{=}0.1$)
then generates ontology-aligned JSON-LD configurations via
protocol-specific prompts and a four-step JSON repair pipeline.
A two-stage abstention gate, combining a reranker-score threshold
and an IRI resolution ratio, prevents unsafe LLM invocations and
filters configurations with insufficient ontology coverage before
SHACL validation. We evaluate every stage against a gold dataset
of 28 field-level queries with evidence-grounded relevance labels:
the hybrid retriever reaches 0.96 HitRate@10 and the cross-encoder
raises MRR@10 from 0.56 to 0.63 while providing perfect
score separation for abstention; the generator attains field-level
$F_1{=}0.87$ with exact-match on 9 of 12 runs; and end-to-end
runs on an H100 GPU complete in 2.6--6.6\,s per device with zero
unsafe writes and zero silent failures on this five-device
proof-of-concept benchmark, every unsuccessful run being flagged
by abstention or deployment verification. The component-wise
evaluation localises the single systematic failure to OPC-UA
JSON-LD generation, a defect invisible to end-to-end metrics
alone. Beyond the synthetic benchmark, a case study commissions a
physics-simulated Universal Robots UR5e collaborative robot
directly from its unmodified vendor documentation (a 254-page
user manual and an 8-page register list, 496 chunks), reaching
field-level $F_1{=}1.0$ over three runs with read-back and
joint-consistency verification. An ablation study and a comparison
with five industrial-LLM systems complete the analysis.
\end{abstract}


% ============================================================
\section{Introduction}
\label{sec:intro}

The proliferation of Industrial Internet of Things (IIoT) devices
in manufacturing, process control, and building automation has
made integration of heterogeneous field devices -- thousands of
sensors, actuators, drives, and PLCs on Modbus RTU, OPC-UA,
PROFIBUS DP, and CANopen -- a critical bottleneck in digital
transformation. Commissioning a device requires a skilled engineer
to locate the manual, identify the correct register map, baud
rate, station address, function codes, and PDO mappings, and
transcribe these into the SCADA/DCS configuration -- a process
that can consume hundreds of engineering hours per cycle and up to
30\% of project cost in greenfield
deployments~\cite{Barnaghi2012semanticsiot}. Manual transcription
is also error-prone: an incorrect register address or baud rate
can silence a sensor or trigger unsafe actuator behaviour.

Recent advances in LLMs~\cite{Brown2020gpt3,OpenAI2023gpt4,Touvron2023llama2}
and retrieval-augmented generation (RAG)~\cite{Lewis2020rag,Gao2024ragsurvey}
open a route to automating this task by grounding generation in
retrieved manual content rather than parametric knowledge alone.
However, three challenges limit direct application in
safety-critical settings: (i)~LLMs hallucinate with high
confidence~\cite{Ji2023hallucination}, and a plausible but invalid
register address can damage field hardware; (ii)~device
descriptions must preserve semantic structure from ECLASS, the
Asset Administration Shell (AAS)~\cite{IEC63278AAS2023}, and
SOSA/SSN~\cite{Janowicz2019sosa} for downstream interoperability;
(iii)~the system must abstain rather than guess when a device is
undocumented, since a silent wrong configuration is more
dangerous than a refusal.

This paper presents \sysname{}, a modular six-stage pipeline
(ingestion, hybrid retrieval, abstention gating, LLM generation,
SHACL validation, protocol deployment) that addresses all three
challenges and is evaluated end-to-end on a four-protocol
benchmark. The contributions are:
\begin{enumerate}
  \item \textbf{Hybrid retrieval with ontology-graph boosting}:
        dense BGE-M3~\cite{Chen2024bgem3} embeddings, BM25, and
        ontology-term boosting fused via Reciprocal Rank
        Fusion~\cite{Cormack2009rrf} and reranked with a
        cross-encoder~\cite{Nogueira2019rerank}.
  \item \textbf{Protocol-specific JSON-LD generation with
        automatic repair}, aligning all output to ECLASS, AAS,
        and SOSA/SSN IRIs.
  \item \textbf{A two-stage, formally motivated abstention gate}
        whose calibration we characterise empirically: retrieval
        scores of documented and undocumented queries separate by
        $8.6\times$, so a wide threshold band achieves perfect
        abstention $F_1$.
  \item \textbf{Four protocol adapters} (Modbus RTU, OPC-UA,
        Profibus DP, CANopen) enforcing deterministic safety
        constraints before any network I/O.
  \item \textbf{A released component-wise evaluation harness}:
        gold data with evidence-grounded relevance labels, and
        separate retriever, generator, and end-to-end evaluations
        with field-level $F_1$, rank metrics, latency breakdown,
        and a retrieval-depth ($k$) sensitivity study, plus
        ablation and security analyses.
\end{enumerate}

% ============================================================
\section{Related Work}
\label{sec:related}

\textbf{RAG and hybrid retrieval.} Lewis~et~al.~\cite{Lewis2020rag}
introduced RAG; Izacard and Grave~\cite{Izacard2021fid} extended
it with Fusion-in-Decoder; Gao~et~al.~\cite{Gao2024ragsurvey}
survey naive, advanced, and modular RAG variants. \sysname{}
extends advanced RAG with ontology-graph boosting, SHACL
validation, and safety-motivated abstention not addressed in
prior surveys. BM25~\cite{Chen2017drqa} remains the standard
sparse baseline; \sysname{} fuses it with dense retrieval via
RRF~\cite{Cormack2009rrf}, adding an ontology-term boost as a
domain-specific third signal. For embeddings, \sysname{} uses
BGE-M3~\cite{Chen2024bgem3} (extending Sentence-BERT-style
encoders~\cite{Reimers2019sbert}) and reranks with a
cross-encoder~\cite{Nogueira2019rerank}.

\textbf{Structured LLM output and abstention.} Few-shot
prompting~\cite{Brown2020gpt3,OpenAI2023gpt4} and open-weight
models such as LLaMA-2~\cite{Touvron2023llama2}, built on the
Transformer~\cite{Vaswani2017attention}, enable structured JSON
generation but still frequently emit malformed JSON below 8B
parameters, motivating \sysname{}'s repair pipeline.
Ji~et~al.~\cite{Ji2023hallucination} taxonomise hallucination as
intrinsic or extrinsic; extrinsic hallucination of protocol
parameters is the primary risk \sysname{} targets. The abstention
gate is motivated by conformal prediction~\cite{Angelopoulos2023conformal},
treating reranker score as a conformity measure, and by LLM
cascade research~\cite{Chen2023frugalgpt} showing that routing
low-confidence queries away from expensive inference reduces cost
without accuracy loss.

\textbf{Ontologies and validation.} SOSA/SSN~\cite{Janowicz2019sosa}
and AAS~\cite{IEC63278AAS2023} provide the vocabulary and submodel
structure for \sysname{}'s JSON-LD~\cite{Sporny2014jsonld} output;
SHACL~\cite{Knublauch2017shacl} provides the formal validation
layer. Barnaghi~et~al.~\cite{Barnaghi2012semanticsiot} motivate
ontology-based IoT interoperability.

\textbf{LLMs in industrial automation.} LLM4PLC~\cite{Faruque2024llm4plc},
Agents4PLC~\cite{Liu2024agents4plc}, and Vendor-Aware
Agents~\cite{Feld2025vendorrag} apply RAG/LLMs to PLC code
generation with compiler or formal-verification feedback, but
none targets fieldbus device configuration, incorporates
ontology-aligned SHACL-validated output, or implements a formally
motivated abstention gate (Table~\ref{tab:comparison}). \sysname{}
is thus the only system combining hybrid retrieval,
ontology-aligned output, formal (SHACL) validation, multi-protocol
deployment, and an abstention gate; the closest prior work,
Vendor-Aware Agents~\cite{Feld2025vendorrag}, uses single-signal
dense retrieval with free-form JSON and no validation or
abstention.

\begin{table}[t]
\centering
\scriptsize
\setlength{\tabcolsep}{2pt}
\caption{Comparison with Related Industrial LLM Systems}
\label{tab:comparison}
\begin{tabular}{@{}lccccc@{}}
\toprule
\textbf{System} & \textbf{Hybrid} & \textbf{Ontol.} & \textbf{Formal} & \textbf{Multi-} & \textbf{Abst.} \\
 & \textbf{Retr.} & \textbf{Out.} & \textbf{Valid.} & \textbf{Proto.} & \textbf{Gate} \\
\midrule
LLM4PLC~\cite{Faruque2024llm4plc}    & \texttimes & \texttimes & Compiler & \texttimes & \texttimes \\
Agents4PLC~\cite{Liu2024agents4plc}  & \texttimes & \texttimes & Formal   & \texttimes & \texttimes \\
Vendor-Aware~\cite{Feld2025vendorrag}& \texttimes & \texttimes & \texttimes & \texttimes & \texttimes \\
Adv.\ RAG~\cite{Gao2024ragsurvey}    & Partial    & \texttimes & \texttimes & \texttimes & Partial \\
\textbf{\sysname{} (this work)}      & \checkmark & \checkmark & SHACL    & \checkmark & \checkmark \\
\bottomrule
\end{tabular}
\end{table}

% ============================================================
\section{System Architecture}
\label{sec:arch}

\begin{figure}[t]
\centering
\begin{tikzpicture}[
  every node/.style={font=\scriptsize},
  st/.style={draw, rounded corners=2pt, fill=blue!10, align=center,
    minimum width=2.2cm, minimum height=0.72cm, inner sep=1pt},
  arr/.style={-{Stealth[length=4pt]}, thick}
]
\node[st] (s1) {\textbf{1 Ingestion}\\ {\scriptsize chunk, BGE-M3,}\\ {\scriptsize BM25, ont.\ graph}};
\node[st, right=0.35cm of s1] (s2) {\textbf{2 Retrieval}\\ {\scriptsize RRF+graph boost,}\\ {\scriptsize rerank top-5}};
\node[st, right=0.35cm of s2] (s3) {\textbf{3 Gate}\\ {\scriptsize $s_{\max}{\geq}\tau$,}\\ {\scriptsize $\rho_{\text{iri}}{\geq}0.80$}};
\node[st, below=0.55cm of s3] (s4) {\textbf{4 LLM gen.}\\ {\scriptsize JSON-LD}\\ {\scriptsize + repair}};
\node[st, left=0.35cm of s4] (s5) {\textbf{5 SHACL}\\ {\scriptsize formal}\\ {\scriptsize validation}};
\node[st, left=0.35cm of s5] (s6) {\textbf{6 Deploy}\\ {\scriptsize adapter +}\\ {\scriptsize read-back}};
\draw[arr] (s1) -- (s2);
\draw[arr] (s2) -- (s3);
\draw[arr] (s3) -- (s4);
\draw[arr] (s4) -- (s5);
\draw[arr] (s5) -- (s6);
\end{tikzpicture}
\caption{End-to-end \sysname{} pipeline. A PDF manual and a
natural-language query enter at Stage~1; a SHACL-valid JSON-LD
configuration is deployed and verified by read-back at Stage~6, or
the pipeline abstains at Stage~3 (pre-LLM) or Stage~4 (post-LLM).}
\label{fig:pipeline}
\end{figure}
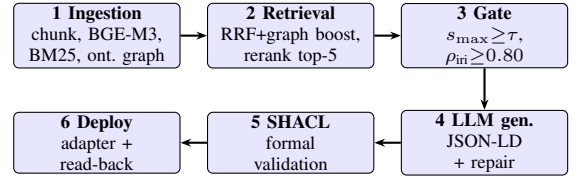

\sysname{} is a six-stage pipeline: ingestion, hybrid retrieval,
abstention gating, LLM generation, SHACL validation, and protocol
deployment, sketched end-to-end in Fig.~\ref{fig:pipeline} and
formalised in Algorithm~\ref{alg:clarity}.

\begin{algorithm}[t]
\caption{\sysname{} End-to-End Pipeline}
\label{alg:clarity}
\begin{algorithmic}[1]
\REQUIRE PDF manual $\mathcal{M}$, query $q$, protocol $p$,
         SHACL shapes $\mathcal{S}_p$, $\tau = 0.72$, $\rho = 0.80$
\ENSURE  Deployed JSON-LD configuration or abstention signal
\STATE \textbf{// Stage 1: Ingestion}
\STATE $\mathcal{C} \leftarrow \texttt{chunk}(\mathcal{M},512,64)$;\;
       $\mathcal{E} \leftarrow \texttt{BGE-M3}(\mathcal{C})$
\STATE $\mathcal{G} \leftarrow \texttt{ontology\_graph}(\text{ECLASS,AAS,SOSA})$
\STATE \textbf{// Stage 2: Hybrid Retrieval}
\STATE $R_{\text{rrf}} \leftarrow \texttt{RRF}(\texttt{dense}(q),
        \texttt{BM25}(q),\,k_{\text{rrf}}{=}60)$; boost matches to $\mathcal{G}$ by $+0.1$
\STATE $R_{\text{top5}} \leftarrow \texttt{CrossEncoder.rerank}(q,R_{\text{rrf}},5)$;\;
        $s_{\max}\leftarrow\sigma(\max r_i.\text{score})$
\STATE \textbf{// Stage 3: Pre-LLM Gate}
\IF{$s_{\max} < \tau$ \textbf{ or } $\rho_{\text{iri}} < 1{-}0.20$}
  \RETURN \textbf{ABSTAIN}
\ENDIF
\STATE \textbf{// Stage 4: LLM Generation + Repair}
\STATE $\hat{y} \leftarrow \texttt{Ollama}(\pi(q,p,R_{\text{top5}}),T{=}0.1)$;\;
       $\hat{y}_{\text{json}} \leftarrow \texttt{JSONRepair}(\hat{y})$
\STATE \textbf{// Stage 5: Post-LLM Gate}
\IF{\texttt{uncertainty\_markers}(\texttt{strip\_json}($\hat{y}$))}
  \RETURN \textbf{ABSTAIN}
\ENDIF
\STATE \textbf{// Stage 6: SHACL Validation + Deployment}
\IF{\textbf{not} \texttt{pyshacl.validate}($\hat{y}_{\text{json}},\mathcal{S}_p$)}
  \RETURN \textbf{INVALID}
\ENDIF
\STATE \texttt{ProtocolAdapter}[$p$].\texttt{deploy}($\hat{y}_{\text{json}}$);\;
       \RETURN $\hat{y}_{\text{json}}$
\end{algorithmic}
\end{algorithm}

\textbf{Ingestion.} PDF manuals are chunked at 512 tokens with
64-token overlap to avoid splitting register-map rows, embedded
with BGE-M3~\cite{Chen2024bgem3} into 1024-d dense vectors stored
in Qdrant (cosine similarity), and indexed in parallel by BM25 for
exact-term matching of register addresses and part numbers. An
ontology graph $\mathcal{G}$ is built from ECLASS, AAS, and
SOSA/SSN labels via \texttt{rdflib}.

\textbf{Hybrid retrieval.} Dense (top-20) and sparse (top-20)
results are fused with RRF (Eq.~\ref{eq:rrf}, constant $k{=}60$),
each chunk receiving a $+0.1$ boost if an ontology label appears
verbatim, before cross-encoder reranking~\cite{Nogueira2019rerank}
to the top-5 context passed to the LLM. Writing $s_i$ for the raw
cross-encoder relevance score of candidate chunk $i$ and $\sigma$
for the logistic sigmoid, the scalar
$s_{\max}=\sigma(\max_i s_i)\in(0,1)$ is the abstention statistic
used in Sec.~\ref{sec:abstention}.
\begin{equation}
  \text{RRF}(d) = \sum_{l \in \{{\rm dense}, {\rm sparse}\}}
    \frac{1}{60 + \text{rank}_l(d)}
  \label{eq:rrf}
\end{equation}

\textbf{Generation and repair.} The LLM (\texttt{llama3.1:8b} via
Ollama, $T{=}0.1$, 2048 tokens) fills a protocol-specific template
with $q$, $R_{\text{top5}}$, and the target JSON-LD
\texttt{@context}. A four-step repair pass (strip JS comments,
remove trailing commas, normalise Python literals, convert
quoting) fixes malformed output before \texttt{json.loads}; at
least one step fired in 11 of 15 benchmark runs.

\textbf{Validation and deployment.} Repaired JSON-LD is parsed to
RDF and validated with \texttt{pyshacl} against protocol-specific
SHACL shapes encoding mandatory properties, datatypes, and value
ranges (e.g. $[1,65534]$ for Modbus registers). Only configurations
passing SHACL reach the protocol adapter.

% ============================================================
\section{Abstention Gate}
\label{sec:abstention}

Let risk event $\mathcal{R}$ be deployment of a configuration
containing at least one extrinsic hallucination. The gate is a
classifier $h:\mathcal{X}\to\{\text{PROCEED},\text{ABSTAIN}\}$
targeting $\Pr[\mathcal{R}\mid h{=}\text{PROCEED}]<\epsilon$,
following conformal prediction~\cite{Angelopoulos2023conformal}
with reranker score as conformity measure.

\textbf{Pre-LLM gate} evaluates two conjunctive criteria: (C1)
$s_{\max}=\sigma(\max_i s_i)\geq\tau{=}0.72$ (Eq.~\ref{eq:pregate});
(C2) IRI resolution ratio $\rho_{\text{iri}}=N_{\text{resolved}}/N_{\text{iri}}\geq0.80$,
where $N_{\text{iri}}$ is the number of ontology IRIs the retrieved
context references and $N_{\text{resolved}}$ the number resolving
against the loaded ECLASS/AAS/SOSA vocabularies.
C1 catches undocumented devices or vague queries; C2 catches
passages using nonstandard terminology the LLM would have to
hallucinate IRIs for.
\begin{equation}
  s_{\max} = \sigma\!\left(\max_{i} s_i\right) \geq \tau=0.72
  \label{eq:pregate}
\end{equation}

\textbf{Post-LLM gate} scans only the prose portion of the LLM
response (outside the JSON block) for uncertainty markers
(``not specified'', ``unknown'', ``I do not know'', etc.);
scanning the full string caused false positives on legitimate
field values such as \texttt{"firmware":"not specified"}.

On the benchmark, the undocumented ``Legacy XYZ'' device stays
far below threshold, so the gate abstains in under 1\,s without
invoking the LLM, while documented device-level queries reach
$s_{\max}\geq0.99$; Section~\ref{sec:eval} calibrates the
threshold empirically over 33 gold queries.

\begin{figure}[t]
\centering
\begin{tikzpicture}[
  node distance=0.32cm and 0pt,
  every node/.style={font=\scriptsize},
  dec/.style={draw, diamond, fill=orange!15, align=center,
    text width=1.7cm, aspect=2.4, inner sep=1pt},
  box/.style={draw, rounded corners=2pt, fill=blue!10,
    minimum width=2.1cm, minimum height=0.5cm, align=center},
  abst/.style={draw, rounded corners=2pt, fill=red!20,
    minimum width=1.7cm, minimum height=0.5cm, align=center,
    font=\scriptsize},
  arr/.style={-{Stealth[length=4pt]}, thick}
]
\node[box] (start) {Top-5 chunks, $s_{\max}$};
\node[dec, below=of start] (d1) {$s_{\max}\!\geq\!0.72$?};
\node[abst, right=0.7cm of d1] (a1) {ABSTAIN\\{\scriptsize score low}};
\node[dec, below=0.4cm of d1] (d2) {$\rho_{\text{iri}}\!\geq\!0.80$?};
\node[abst, right=0.7cm of d2] (a2) {ABSTAIN\\{\scriptsize IRI gap}};
\node[box, below=0.4cm of d2] (llm) {LLM gen.\ + repair};
\node[dec, below=0.4cm of llm] (d3) {Uncertainty\\marker?};
\node[abst, right=0.7cm of d3] (a3) {ABSTAIN\\{\scriptsize uncertain}};
\node[box, below=0.45cm of d3] (shacl) {SHACL + deploy};
\draw[arr] (start) -- (d1);
\draw[arr] (d1) -- node[left,font=\tiny]{yes} (d2);
\draw[arr] (d1) -- node[above,font=\tiny]{no} (a1);
\draw[arr] (d2) -- node[left,font=\tiny]{yes} (llm);
\draw[arr] (d2) -- node[above,font=\tiny]{no} (a2);
\draw[arr] (llm) -- (d3);
\draw[arr] (d3) -- node[left,font=\tiny]{no} (shacl);
\draw[arr] (d3) -- node[above,font=\tiny]{yes} (a3);
\end{tikzpicture}
\caption{Two-stage abstention gate. The pre-LLM gate (top two
diamonds) fires before inference; the post-LLM gate inspects only
the prose portion of the LLM response.}
\label{fig:abstention}
\end{figure}
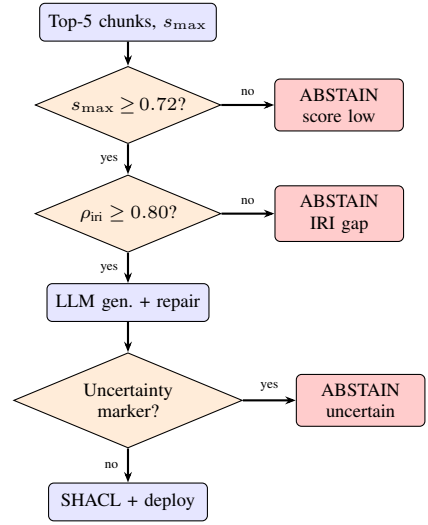

% ============================================================
\section{Protocol Adapters}
\label{sec:adapters}

Each of the four adapters enforces a deterministic safety layer
before any network I/O, deploys validated JSON-LD against a live
protocol simulator (\texttt{pymodbus} on :5020, \texttt{asyncua}
on :4840) or structural simulation for Profibus/CANopen, and
verifies every written value by read-back. Hard constraints
include Modbus register addresses in $[1,65534]$ with function
codes in $\{1,2,3,4,5,6,15,16\}$ (0 and 65535 permanently
blocked), OPC-UA Node-ID syntax and datatype checks, Profibus
station addresses in $[0,125]$, and CANopen node IDs in $[1,127]$
with PDO format and range checks.

% ============================================================
\section{Evaluation}
\label{sec:eval}

Industrial adoption requires component-wise, reproducible
evidence, not a single end-to-end demonstration. We therefore
evaluate the retriever, the generator, and the full pipeline
separately against gold data, and release the harness with the
code.

\subsection{Methodology and Gold Dataset}
The corpus is produced by the pipeline's own chunker (512-token
windows, 64 overlap) over the four benchmark manuals plus a parsed
8-page PDF: 71 chunks, with deliberate cross-device confusability
(three devices specify baud/bit rates, two expose node IDs). The
gold set has 28 field-level queries (7 per device) and 5
out-of-corpus queries expected to abstain. Relevance labels are
\emph{evidence-grounded}: a chunk is relevant iff it contains the
exact manual statement of the gold value, making labels
reproducible rather than judgement-based. Each device carries a
gold configuration (3--6 mandatory fields) for generator scoring.
Retrieval runs on a CPU (Xeon) node; generation uses
\texttt{llama3.1:8b} on Ollama on one NVIDIA H100, three runs per
device at $T{=}0.1$.

\subsection{Retriever Evaluation}
Table~\ref{tab:retrieval} compares the four retrieval
configurations. Hybrid RRF fusion achieves the best coverage
(HitRate@10 $0.964$ vs.\ $0.929$ for either signal alone),
confirming that dense and sparse retrieval fail on
\emph{different} queries: BM25 wins exact identifiers
(\texttt{P0918}, \texttt{0x18A}) while dense embeddings win
paraphrases. The cross-encoder improves early precision -- MRR@10
rises from $0.557$ to $0.625$, nDCG@10 from $0.634$ to $0.679$,
and the mean first-relevant rank drops from $3.50$ to $3.18$
(rank~27 to rank~9 in the extreme case). Reranking is not
uniformly positive (one query's chunk left the top-5, HitRate@5
$0.893\!\to\!0.857$), but its decisive benefit is calibration: raw
RRF and BM25 scores are unbounded and corpus-dependent, whereas
the sigmoid cross-encoder score supports the fixed abstention
threshold evaluated next. Retrieval is cheap: embedding 59\,ms,
search ${<}1$\,ms, reranking 20 candidates 321\,ms on CPU.

\begin{table}[t]
\centering
\small
\setlength{\tabcolsep}{2.5pt}
\caption{Retriever Evaluation (28 gold queries, 71 chunks; MRR/nDCG at cutoff 10)}
\label{tab:retrieval}
\begin{tabular}{@{}lccccc@{}}
\toprule
\textbf{Variant} & \textbf{HR@1} & \textbf{HR@5} & \textbf{HR@10} & \textbf{MRR} & \textbf{nDCG} \\
\midrule
Dense (BGE-M3)      & 0.357 & 0.857 & 0.929 & 0.524 & 0.561 \\
Sparse (BM25)       & 0.429 & 0.857 & 0.929 & 0.623 & 0.647 \\
Hybrid (RRF+graph)  & 0.321 & 0.893 & \textbf{0.964} & 0.557 & 0.634 \\
~+ Cross-encoder    & \textbf{0.429} & 0.857 & \textbf{0.964} & \textbf{0.625} & \textbf{0.679} \\
\bottomrule
\end{tabular}
\end{table}

\subsection{Abstention-Gate Calibration}
On the gold set, every documented query reaches
$s_{\max}\geq0.526$ and every out-of-corpus query stays at
$s_{\max}\leq0.061$ -- an $8.6\times$ separation with no overlap,
so any threshold $\tau\in(0.061,0.526)$ attains abstention
precision~$=$~recall~$=1.0$. The production threshold
$\tau{=}0.72$ sits above this band: on hard field-level queries
it falsely abstains once (precision $0.833$, recall $1.0$), which
errs in the safe direction, and on the device-level commissioning
queries used by the pipeline ($s_{\max}\geq0.99$) it causes no
false abstention, giving operators an empirical basis for tuning
$\tau$.

\subsection{Generator Evaluation}
Table~\ref{tab:generator} scores generated JSON-LD against gold
fields (a field is correct iff its key appears anywhere in the
output with the normalised gold value). Micro-averaged over 12
runs: precision $0.830$, recall $0.907$, $F_1$ $0.867$,
exact-match 9/12. Modbus, Profibus, and CANopen are perfect in
all runs; OPC-UA fails in all three. The component-wise design
localises the fault precisely: the prompt demonstrably contained
the required node identifiers (five occurrences of
\texttt{ns=2;i=1001}), yet \texttt{llama3.1:8b} emitted only the
server endpoint and omitted the mandated \texttt{clarity:nodes}
array -- a model limitation on the most deeply nested schema, not
a retrieval failure. The OPC-UA SHACL shape does not yet mandate a
non-empty node array (a gap this evaluation exposed), so the
incomplete configuration was caught one layer later by deployment
verification. No incorrect value was ever written to a device.

\begin{table}[t]
\centering
\small
\setlength{\tabcolsep}{4pt}
\caption{Generator Field-Level Scores (3 runs/device)}
\label{tab:generator}
\begin{tabular}{@{}lcccc@{}}
\toprule
\textbf{Protocol} & \textbf{Prec.} & \textbf{Rec.} & \textbf{$F_1$} & \textbf{Exact match} \\
\midrule
Modbus RTU   & 1.000 & 1.000 & 1.000 & 3/3 \\
OPC-UA       & 0.000 & 0.000 & 0.000 & 0/3 \\
Profibus DP  & 1.000 & 1.000 & 1.000 & 3/3 \\
CANopen      & 1.000 & 1.000 & 1.000 & 3/3 \\
\midrule
\textbf{Micro avg.} & 0.830 & 0.907 & 0.867 & 9/12 \\
\bottomrule
\end{tabular}
\end{table}

\subsection{End-to-End Results}
Table~\ref{tab:e2e} reports full-pipeline outcomes over 15 runs.
All Modbus, Profibus, and CANopen deployments succeed except one
VFD300 run in which the post-LLM gate falsely abstained on an
uncertainty marker; OPC-UA never deploys, for the generator reason
above; the undocumented device abstains in 0.5\,s (pre-LLM gate,
no inference cost). Crucially for industrial trust, \emph{failure
detection is 100\%}: every unsuccessful run terminated in an
explicit abstention or a deployment-verification error, never a
silently wrong configuration, and zero safety-constraint
violations occurred. All configurations passed SHACL (validity
rate 1.0); end-to-end abstention precision/recall is $0.75/1.0$.

\begin{table}[t]
\centering
\small
\setlength{\tabcolsep}{3pt}
\caption{End-to-End Results on H100 (3 runs/device)}
\label{tab:e2e}
\begin{tabular}{@{}llccc@{}}
\toprule
\textbf{Device} & \textbf{Protocol} & \textbf{Deploy} & \textbf{Abstain} & \textbf{Lat.\ (s)} \\
\midrule
TH200 Temp.  & Modbus RTU  & 3/3 & 0/3 & 6.5 \\
PS500 Press. & OPC-UA      & 0/3$^{\dagger}$ & 0/3 & 3.6 \\
VFD300 Drive & Profibus DP & 2/3 & 1/3$^{\ddagger}$ & 3.4 \\
ENC100 Enc.  & CANopen     & 3/3 & 0/3 & 3.1 \\
Legacy XYZ   & Modbus RTU  & 0/3 & 3/3 (correct) & 0.5 \\
\bottomrule
\multicolumn{5}{@{}p{6.8cm}@{}}{\scriptsize $^{\dagger}$Incomplete
generation detected by deployment verification (Sec.~VI-D).
$^{\ddagger}$False abstention by the post-LLM uncertainty-marker
gate.}
\end{tabular}
\end{table}

\subsection{Latency and Retrieval Depth}
Mean per-stage latency on the H100: query embedding 248\,ms, dense
$+$ BM25 search ${<}2$\,ms, reranking 449\,ms, LLM generation
4.69\,s ($\approx$87\%), SHACL 10\,ms, deployment 29\,ms --
documented devices complete in 2.6--6.6\,s, versus 35--93\,min on
CPU-only inference (0.8 vs.\ ${\sim}120$ tokens/s), a
two-to-three-orders-of-magnitude reduction that makes interactive
commissioning practical. Because the LLM dominates, prompt size
(retrieval depth $k$) is the main lever: sweeping
$k\in\{1,3,5,10\}$ on TH200, latency triples ($2.4$ to $7.3$\,s)
while field $F_1$ is non-monotonic -- at $k{=}3$ a distractor
chunk displaced a needed one and cost a field ($F_1$ $0.667$),
whereas $k{=}5$ holds the evidence in a $\sim$1.4k-token prompt at
$F_1$ $1.0$. Passing many low-ranked chunks is thus both slower
\emph{and} riskier; the pre-LLM gate adds the complementary saving
of 0.5\,s per undocumented query instead of a full round-trip.

\subsection{Ablation}
Table~\ref{tab:ablation} isolates each component by disabling it
while holding the rest fixed (one run per configuration on the CPU
host, in a session where all four documented devices deployed).
Removing BM25 fails VFD300, whose station-address field is
retrievable only by exact-term matching; removing dense retrieval
fails two devices where BM25 cannot bridge the semantic query
manual gap -- both consistent with the per-signal results in
Table~\ref{tab:retrieval}. Removing the pre-LLM gate is a cost
failure: the undocumented device consumes a full LLM round-trip
before rejection (73~min on the CPU host; ${\sim}7$\,s on the
H100) instead of abstaining in under a second. Removing the
four-step JSON repair drops deployment from 4/4 to 1/4, confirming
that sub-8B models often emit malformed JSON that a lightweight
deterministic repair recovers without a second LLM call. Removing
SHACL still deploys here but forfeits formal verification: a
reserved register address of 0 would pass undetected to the field,
precisely the silent error the system exists to prevent.

\begin{table}[t]
\centering
\small
\setlength{\tabcolsep}{4pt}
\caption{Ablation: Deployment Success (Single CPU-Host Run/Config.)}
\label{tab:ablation}
\begin{tabular}{@{}lcp{4.0cm}@{}}
\toprule
\textbf{Configuration} & \textbf{Dep.} & \textbf{Failure mode when removed} \\
\midrule
Full \sysname{}      & 4/4 & --- \\
$-$ BM25 (sparse)    & 3/4 & Exact-ID field unretrievable \\
$-$ Dense retrieval  & 2/4 & Semantic query--manual gap \\
$-$ Pre-LLM gate     & 4/4 & +73~min wasted on undocumented \\
$-$ JSON repair      & 1/4 & Malformed JSON unparsed \\
$-$ SHACL validation & 4/4 & No formal constraint guarantee \\
\bottomrule
\end{tabular}
\end{table}

\subsection{Case Study: Unmodified Vendor Documentation}
\label{sec:ur5e}
To probe generalisation beyond the synthetic benchmark, we ran the
unchanged pipeline on a Universal Robots UR5e collaborative robot.
Its knowledge base is the real, unaltered vendor documentation --
a 254-page user manual and an 8-page Modbus register list (496
chunks total, no manual re-formatting) -- and its actuation is a
MuJoCo physics simulation exposed through the robot's documented
Modbus TCP server, so every deployed write and read-back is
exercised against a moving robot. From the natural-language
request \emph{``initialise the PLC data-exchange handshake: write
0,0,0,1 to holding registers 128--131 using function code 6
(UINT16)''}, \sysname{} generated and deployed the correct
configuration in all three runs (field-level $F_1{=}1.0$, exact
match) in 4.2 and 4.7\,s per warm run (14.0\,s including one-off
model load). A sentinel preload (60000) confirmed each write
originated from \sysname{}; post-deployment read-back returned the
gold values ($128{:}0,129{:}0,130{:}0,131{:}1$); the robot-mode
register (258) read 7 (Running); and across 18 joint-angle
comparisons the milliradian-encoded joint registers matched the
simulator ground truth within tolerance, values changing between
reads taken 1\,s apart (liveness). No SHACL or safety-constraint
violation occurred.

The real manual also stresses retrieval and calibration. Hybrid
HitRate@10 falls to $0.40$ on the layout-heavy 254-page manual,
and the documented/undocumented $s_{\max}$ separation narrows to
$[0.070,0.156]$, so the production $\tau{=}0.72$ would falsely
abstain on five valid field queries; recalibrating $\tau$ to the
separation midpoint ($0.113$) restores correct decisions on all 12
gold queries. This is direct evidence for the per-corpus threshold
tuning the sweep prescribes, and for the layout-aware ingestion
larger brownfield catalogues will need.

% ============================================================
\section{Security Analysis}
\label{sec:security}

Industrial RAG systems face \emph{manual-content prompt
injection}: adversarial text in ingested chunks that attempts to
steer the LLM into unsafe output. For an attacker with write
access to one or more indexed chunks, we consider three vectors --
\textbf{A1} direct instruction override; \textbf{A2}
plausible-but-incorrect parameter values planted for the LLM to
copy; \textbf{A3} retrieval-score manipulation via keyword
stuffing. \sysname{} layers defences against all three: the SHACL
and protocol-adapter constraints reject safety-violating output
regardless of LLM instruction-following (A1); the reranker
penalises incoherent stuffed text and SHACL bounds catch
out-of-range values (A2); the bounded $+0.1$ graph boost cannot
dominate reranker relevance (A3). The main residual risk is a
plausible in-range value that both reranks highly and evades SHACL
shapes -- to be mitigated by document signing and chunk-level
provenance attestation; \sysname{} already logs a
\texttt{provenance\_trail} (document, page, confidence) per chunk
for IEC~62443 auditability.

% ============================================================
\section{Discussion and Conclusion}
\label{sec:discussion}

The component-wise evaluation surfaced two concrete defects that
end-to-end metrics alone would have conflated: a
\texttt{llama3.1:8b} generation weakness on the deeply nested
OPC-UA schema (fixable by schema-constrained decoding or few-shot
node examples) and an OPC-UA SHACL shape that must additionally
mandate a non-empty node array. The core benchmark remains synthetic and small; the
UR5e case study (Sec.~\ref{sec:ur5e}) is a first step onto real
vendor documentation, but scaling to full brownfield catalogues
with multi-column tables and scanned pages will require stronger
document-layout extraction (e.g.\ LayoutLM~\cite{Xu2020layoutlm})
and a larger gold set. The threshold sweep gives an empirical band
for $\tau$, and the UR5e run shows it must be recalibrated
per corpus; a conformal calibration on a held-out set would add a
formal $1{-}\alpha$ coverage
guarantee~\cite{Angelopoulos2023conformal}.

Two deployment concerns remain open. When multiple manual
revisions disagree, \sysname{} currently ranks purely by
relevance; the logged \texttt{provenance\_trail} (document, page,
confidence) already carries the metadata needed to add a
revision- or reliability-weighted prior to RRF, so a newer or
vendor-authoritative document outranks a stale one. Extension is
additive: new devices, manuals, or ontology revisions are onboarded
by re-chunking and re-embedding the affected documents and
reloading the ECLASS/AAS/SOSA graph, with no retraining, so
post-deployment maintenance is an ingestion operation rather than
a model update. Future work includes quantisation benchmarking,
brownfield evaluation, additional protocol adapters (HART~7,
EtherNet/IP, IO-Link), and AAS Technical Data submodel
integration.

This paper presented \sysname{}, a production-oriented pipeline
combining hybrid ontology-boosted retrieval, LLM generation with
automatic JSON repair, a two-stage abstention gate, SHACL
validation, and multi-protocol deployment adapters, evaluated
component-wise against gold data. The hybrid retriever reaches
HitRate@10 $0.964$ (MRR@10 $0.625$ with reranking); the generator
attains field-level $F_1{=}0.867$ (perfect on three of four
protocols); and the full pipeline commissions documented devices
in 2.6--6.6\,s on a single H100 with zero unsafe writes and 100\%
failure detection across all 15 benchmark runs, and commissions a
UR5e from unmodified vendor documentation. By grounding generation
in retrieved evidence, abstaining when that evidence is
insufficient, and verifying every deployment, \sysname{} provides
the transparent, reproducible safety case that industrial adoption
demands.

% ============================================================
\bibliographystyle{IEEEtran}
\bibliography{references}

\end{document}